\documentclass{svproc}
\usepackage{url}

\usepackage{color}
\usepackage{graphicx}
\usepackage[square, numbers]{natbib}

\definecolor{turquoise}{cmyk}{0.65,0,0.1,0.3}
\definecolor{purple}{rgb}{0.65,0,0.65}
\definecolor{dark_green}{rgb}{0, 0.5, 0}
\definecolor{orange}{rgb}{0.8, 0.6, 0.2}
\definecolor{red}{rgb}{0.8, 0.2, 0.2}
\definecolor{darkred}{rgb}{0.6, 0.1, 0.05}
\definecolor{blueish}{rgb}{0.0, 0.3, .6}
\definecolor{light_gray}{rgb}{0.7, 0.7, .7}
\definecolor{pink}{rgb}{1, 0, 1}
\definecolor{greyblue}{rgb}{0.25, 0.25, 1}

\begin{document}
\mainmatter              
\title{Exploiting CLIP-based Multi-modal Approach for Artwork Classification and Retrieval}
\titlerunning{Exploiting CLIP-based Multi-modal Approach for Artwork Classification}  
%
\author{Alberto Baldrati\inst{1,2} \and  Marco Bertini\inst{1} \and Tiberio Uricchio\inst{1} \and \\ Alberto Del Bimbo\inst{1}}
\authorrunning{Baldrati et al.} 
%
%
\institute{Universit\`a degli Studi di Firenze - MICC,\\
\email{[name.surname]@unifi.it}\\
Firenze, Italy
\and
Universit\`a di Pisa \\
Pisa, Italy}
\maketitle              

\begin{abstract}
Given the recent advantages in multimodal image pretraining where visual models trained with semantically dense textual supervision tend to have better generalization capabilities than those trained using categorical attributes or through unsupervised techniques, in this work we investigate how recent CLIP model can be applied in several tasks in artwork domain.
We perform exhaustive experiments on the NoisyArt dataset which is a collection of artwork images collected from public resources on the web.
On such dataset CLIP achieve impressive results on (zero-shot) classification and promising results in both artwork-to-artwork and description-to-artwork domain.

\keywords{image retrieval, zero-shot classification, artwork, CLIP}
\end{abstract}
\section{Introduction}
Image Classification and Content-Based Image Retrieval (CBIR) are fundamental tasks for many domains, and have been thoroughly studied by the multimedia and computer vision communities. In cultural heritage domain, these tasks allow to simplify the management of large collections of images, allowing to annotate, search and explore them more easily and with lower costs.

In the latest years neural networks have proved to outperform engineered features in both tasks. These networks are typically used in an unimodal fashion, i.e.~only one media is used to train and use a network. This may limit the types of application that can be developed and may also reduce the performance of the networks. Several recent works are showing how using multi-modal approaches may improve the performance in several tasks related to visual information.
In \cite{radford2021learning} it has been shown that CLIP, a model trained using an image-caption objective alignment on a giant dataset made of 400 million (image, text) pairs, obtain impressive results on several downstream tasks. The authors pointed out that, using only textual supervision, CLIP model learns to perform a wide set of tasks during pre-training including OCR, geo-localization, action recognition and many others. This task learning can be leveraged via natural language prompting to enable zero-shot transfer to many existing dataset.

In this work we try to exploit the zero-shot capabilities of CLIP in the artworks domain, in particular we focus on the NoisyArt \cite{del2019noisyart} dataset which is originally designed to support research on webly-supervised recognition of artworks and Zero-Shot Learning (ZSL). Webly-supervised learning is interesting since it allows to greatly reduce annotation costs required to train deep neural networks, thus allowing cultural institutions to train and develop deep learning methods while keeping their budgets for the curation of their collections rather than the curation of training datasets. In Zero-Shot Learning approaches visual categories are acquired without any training samples, exploiting the alignment of semantic and visual information learned on some training dataset. ZSL in artwork recognition is a problem of instance recognition, unlike the other common ZSL problems that address class recognition. Zero-shot recognition is particularly appealing for cultural heritage and artwork recognition, although it is an extremely challenging problem, since it can be reasonably expected that museums have a set of curated description paired with artworks in their collections.

To get a better idea of how CLIP behaves in the artworks domain we started with a classification task using a shallow classifier and CLIP as the backbone.
Subsequently, thanks to the descriptions of the artworks in the dataset, we performed experiments in the field of zero-shot classification where CLIP was able to demonstrate its abilities in this task.
Finally, we performed experiments on the tasks of artwork-to-artwork and description-to-artwork retrieval obtaining very promising results and superior performance to a ResNet-50 pre-trained on ImageNet \cite{russakovsky2015imagenet}

\section{Related Works}
Regarding CBIR, after the introduction of the successful Bag-of-Visual-Words model in \cite{sivic-2003} that use engineered visual features such as SIFT points, many works have improved the performance addressing different aspects such as approximating local descriptors \cite{Jegou-2010}, learning improved codebooks \cite{mikulik-2013}, improving local features aggregation \cite{perronnin-2010, jegou-2012, delhumeau2013revisiting}. In the last years, following the success obtained using Convolutional Neural Networks (CNN) to address the problem of image classification \cite{krizhevsky2012imagenet}, CNN-based features have started to be used also for image retrieval tasks. A complete survey that compares SIFT-based and CNN-based methods for instance-based image retrieval is presented in \cite{zheng2017sift}.
Commonly used backbone networks are VGG \cite{simonyan2014very} and ResNet \cite{Kaiming-2016}, typically pretrained on ImageNet and then fine tuned for a specific domain.
CNN features have been pooled using techniques like Regional maximum activation of convolutions (R-MAC) \cite{tolias2016particular}. R-MAC considers a set of fixed squared regions at different scales, collecting the maximum response in each channel and then sum-pooling them to create the final R-MAC descriptor.
More recent works follow an end-to-end approach: in \cite{arandjelovic2016netvlad} has been proposed a layer called NetVLAD, pluggable in any CNN architecture and trainable through back-propagation, that allows to train end-to-end a network using an aggregation of VGG16 convolutional activations. Multi-scale pooling of CNN features followed by NetVLAD has been proposed in \cite{vaccaro2020image}, obtaining state-of-the-art results using VGG16. In \cite{radenovic2018fine} a trainable pooling layer called  Generalized-Mean (GeM) has been proposed, along with learning whitening, for short representations. In this work a two stream Siamese network is trained using contrastive loss. The authors use up to 5 image scales to extract features.

\section{Dataset}
NoisyArt \cite{del2019noisyart} is a collection of artwork images collected using articulated queries to metadata repositories and image search engines on the web. According to the creators of the dataset, the goal of NoisyArt is to support research on webly-supervised artwork recognition for cultural heritage applications.

In Table \ref{tab:noisyart} the characteristics of the NoisyArt dataset are summarized
\begin{table}[ht!]
\centering
\begin{tabular}{l r r r r}
&&\multicolumn{2}{c}{\textbf{(webly images)}} & \textbf{(verified images)}\\
&\textbf{classes} & \textbf{training} & \textbf{validation} & \textbf{test}\\

\hline
& 2,920 & 65,759 & 17,368 & 0  \\
& 200 & 4,715 & 1,253 & 1,355 \\ \hline
\textbf{totals} & 3,120 & 70,474 & 18,621 & 1,355\\
\end{tabular}

\caption{Characteristics of the NoisyArt dataset}
\label{tab:noisyart}
\end{table}

NoisyArt is a complex dataset which can be used on a wide variety of automated recognition problems. The dataset is particularly well suited to webly supervised instance recognition as a weakly-supervised extension of fully-supervised learning. 
In the dataset, for testing purposes, a subset of classes with manually verified test images is provided (\textit{i.e.} with no label noise).

\begin{figure}[htb]
    \centering
    \includegraphics[width=\linewidth]{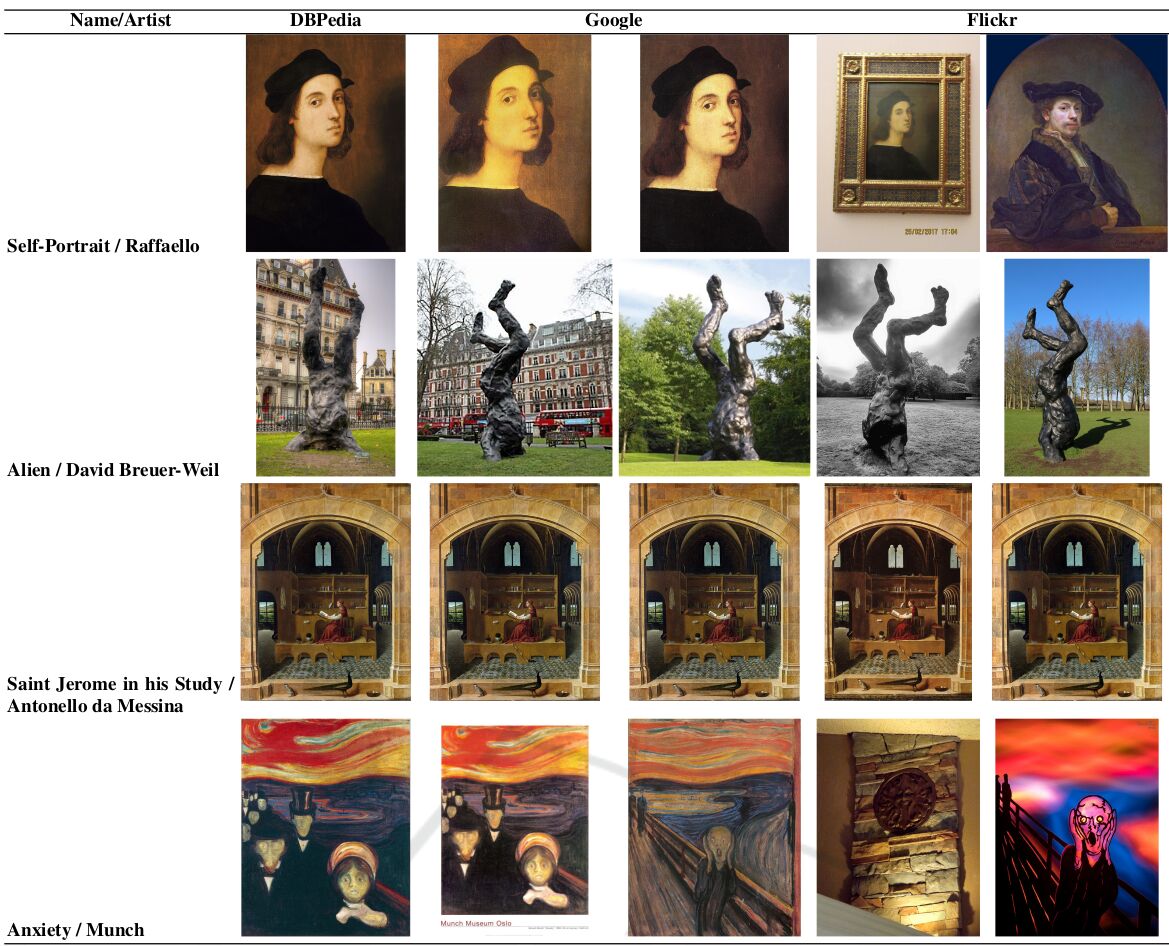}
    \caption{Sample classes and training images from the NoisyArt dataset. For each artwork/artist pair we show the seed image obtained from DBpedia, the first two Google Image search results, and first two Flickr search. Image taken from \cite{del2019noisyart}}
    \label{fig:noisyart}
\end{figure}

The NoisyArt dataset is collected from numerous public resources available on the web. These resources are DBpedia (where also the metadata are retrieved), Google Images and Flickr.
Figure \ref{fig:noisyart} shows some examples of artworks with their respective sources.

From these sources the authors collected 89,095 images divided into 3,120 classes. Each class contains a minimum of 20 images and a maximum of 33. To make sure to have a non-noisy and more reliable test set the authors decided to create a supervised test set using a small subset of the original classes: 200 classes containing more than 1,300 images taken from the web or from personal photos. This test set is not balanced: for some classes we have few images, and some others have up to 12.
The different method of collecting training and test sets also raises the issue of a strong domain shift between these images and those in the training set.
Finally, each artwork has a description and metadata retrieved from DBpedia, from which a single textual document was created for each class. These descriptions are included in the dataset to support research on zero-shot learning and other multi-modal approaches to learning over weakly supervised data.

\subsection{GradCAM visualization}\label{sec:noisy-gradcam}
In order to have a better idea of the portions of the image that CLIP considers most important when it associates a text with an image, before moving on to the quantitative experiments, we carried out some qualitative tests using the well-know visualization technique gradCAM \cite{Selvaraju2019gradcam}.
The technique we used is a generalization of gradCAM, where, instead of computing gradients with respect to an output class, gradients are computed with respect to textual features computed with CLIP's text encoder from the description.
This approach makes each heat-map calculated by gradCAM dependent on the individual description, showing us the portions of the image that CLIP most closely associates with it.
As a common practice the \textit{saliency layer} used is the last convolutional layer of CLIP's visual encoder.

\begin{figure}[ht!]
    \centering
    \includegraphics[width=0.48\linewidth]{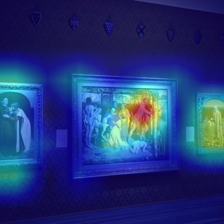}
    \vspace{2.2ex}
    \hfill
    \includegraphics[width=0.48\linewidth]{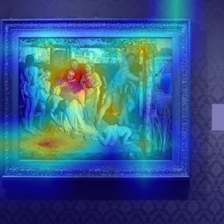}
    \includegraphics[width=0.48\linewidth]{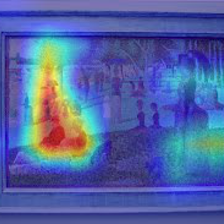}
    \hfill
    \includegraphics[width=0.48\linewidth]{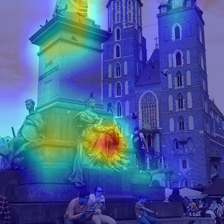}
    \caption{Examples of gradCAM visualization on NoisyArt computing the gradients with respect to the description CLIP text features}
    \label{fig:gradCAM}
\end{figure}

Figure \ref{fig:gradCAM} shows four examples of gradCAM visualization. We can see how, using the descriptions in the dataset, CLIP places attention to the most significant portions of the image. This fact made us confident that CLIP would work very well in the domain of artwork.

\section{Experiments}
\subsection{Webly-supervised Classification}
To test the performance of CLIP in the art domain, following the experimental setup followed by the authors of the dataset, we performed a webly-supervised classification on the 200 classes that are also available in the test set.

\subsubsection{Experimental Setup}
Given an input image $\mathbf{x}$, we extract a feature vector using only the CLIP image encoder and then we pass it through a shallow classifier, consisting of a single hidden layer and an output layer that estimates class probabilities $p(c\mid \mathbf{x})$. The hidden layer is followed by an  $L^2$-normalization layer which, as noted in \cite{delchiaro2019420}, helps to create similar representations for image with different visual characteristics because the magnitude of features is ignored by the final classification layer. Such normalization is therefore useful to alleviate the effects of the domain shift between training and test set.

The structure of the shallow classifier is basically the same of \cite{del2019noisyart, kalantidis2016crossdimensional}. This choice was made intentionally to analyze the effects of using the CLIP image encoder instead of a convolutional backbone trained on ImageNet.
For mitigating and identifying label noise during training in \cite{del2019noisyart} several techniques like Labelflip noise, entropy scaling for Outlier Mitigation and Gradual Bootstrapping are used. In our experiments however, following \cite{delchiaro2019420}, we only use the $L^2$-normalization layer after the hidden layer.

We trained such shallow classifier for 300 epochs with a batch size of 64, the learning rate used was $1e-4$. We used the CLIP model which has as convolutional backbone a slightly modified version of the ResNet-50. The hidden layer has an input dimension of 1024 (CLIP output dimension) and output dimension of 4096.

\subsubsection{Experimental Results}
\begin{table}[htb]
\centering
\begin{tabular}{c|| c|c |c|c}
&\multicolumn{2}{c}{\textbf{test}}  &\multicolumn{2}{c}{\textbf{validation}} \\
\textbf{Model} & \textbf{acc} & \textbf{mAP} & \textbf{acc} & \textbf{mAP}\\
\hline\hline
\textbf{RN50 BL \cite{del2019noisyart}} & 64.80 & 51.69 & 76.14 & 63.08\\ \hline
\textbf{RN50 BS \cite{del2019noisyart}} & 68.27 & 57.44 & 75.98 & 62.83\\ \hline
\textbf{RN50 $\mathbf{\alpha=0.4}$ \cite{delchiaro2019420}} & 74.89 & 62.86 & 77.14 & 63.71 \\ \hline\hline
\textbf{CLIP RN50} & \textbf{86.63} & \textbf{77.88} & \textbf{83.56} & \textbf{72.23} \\ \hline
\end{tabular}
\caption{Recognition accuracy (acc) and mean Average Precision (mAP) on NoisyArt dataset}
\label{tab:noisyart-class}
\end{table}

Table \ref{tab:noisyart-class} summarizes the experimental results we obtained in this classification setting. In the table \textit{BL} refers to the baseline network \cite{del2019noisyart} without any sort of label mitigating approach, \textit{BS} refers to the noisy mitigating approach of \cite{del2019noisyart} and \textit{RN50 $\alpha=0.4$} refers to the normalization approach of \cite{delchiaro2019420} where the $L^2$-normalization is scaled by $\alpha$.

From the table it is immediately evident that with the use of CLIP as a backbone it is possible to obtain very significant improvements both on the test and the validation set. It is very interesting to see that \cite{del2019noisyart, delchiaro2019420} have better results on validation than on the test set. In our case, however, the situation is reversed by having comparable and slightly better results on the test set. This demonstrates how CLIP is quite robust to domain shift being it able to extract the semantic of an image regardless of its raw content.

\subsection{Zero-shot Classification}\label{zeroshot-noisy}
The availability of descriptions associated with artwork made it possible to perform experiments in the area of zero-shot classification by exploiting CLIP's ability to assign a similarity score between text and images.

\subsubsection{Experimental Results}
\begin{table}[htb]
\centering
\begin{tabular}{c|| c|c}
\textbf{Model} & \textbf{acc} & \textbf{mAP}\\
\hline\hline
\textbf{DEVISE RN50 \cite{frome2013devise}} & 24.79 & 31.90\\ \hline
\textbf{EsZSL RN50 \cite{paredes2015simple}} & 25.63 & 29.89\\ \hline
\textbf{COS+NLL+L2 RN50 \cite{delchiaro2019420}} & 34.93 & 45.53 \\ \hline\hline
\textbf{CLIP RN50} & \textbf{60.27} & \textbf{69.23}\\ \hline

\end{tabular}
\caption{Zero-shot recognition accuracy (acc) and mean Average Precision (mAP) on NoisyArt dataset}
\label{tab:noisyart-zeroclass}
\end{table}

Table \ref{tab:noisyart-zeroclass} shows the immense potential in the zero-shot classification domain of CLIP. As a matter of fact, comparing the results with those found in the literature, we notice that by using CLIP, improvements of over 20\% can be achieved.
It is also worth noting that the results we have compared ours with have been achieved through a training process that uses a three-fold cross validation where the 200 verified classes are divided into 150 for training/validation and 50 for zero-shot test classes. On our side we used CLIP out-of-the-box without any training on NoisyArt dataset.

In order to make a complete argument, it is also necessary to mention that since the data on which CLIP was actually trained is not public, we do not know if any images from this dataset were used in its training process. If so we would have some sort of leak of information that would make the comparison less fair.

\subsection{Image Retrieval}
Seeing the excellent behavior of CLIP in the (zero-shot) classification of artwork, we decided to perform some experiments in image retrieval.

In all the experiments that we are going to present, the images contained in the validation set (1253 images belonging to the 200 verified classes) were used as queries, while those of the test set (1379 images of the same 200 classes) were used as index images.

\subsubsection{Experimental Setup}
We have conducted numerous experiments to make sure that we have a complete idea of how CLIP performs in this task on the NoisyArt dataset.
As in classification experiments the CLIP model which has as visual backbone a modified version of the ResNet-50 is used.

The most natural way to use CLIP in retrieval is obviously to use the output of the visual encoder as global descriptor comparing only the visual features, that is exactly what we did initially as a first experiment.

To take advantage of the CLIP textual encoder and of its goodness in zero-shot classification, we then reinterpreted the image-to-image retrieval task as zero-shot classification followed by text-to-image retrieval.
This reinterpretation was made possible by the description and the metadata associated to each class.
Thus given a query, zero-shot classification of that image was performed as first phase, by exploiting CLIP's ability to link images and texts. We therefore used CLIP to assign a similarity score to each possible (query image, artwork description) pair using the description with the highest score in the second phase.

\noindent The second phase consists in comparing the description chosen at the end of the first one with all the images in the dataset, assigning a similarity score to each possible (query description, index image).
For a complete comparison in the results we have also reported an experiment where the first part of classification is bypassed and the correct monument is always used in the text-to-image retrieval phase.

Another setup we experimented with consists in adding to the zero-shot classification followed by text-to-image retrieval experiment as a re-ranking phase where the first 100 retrieved images are re-ordered using the similarity of the visual features.

Finally, the CLIP network was fine-tuned for adapting to this task. The fine-tuning process was done by inserting a shallow classifier composed of two linear layers at the output of the visual encoder. The learning rate was set to $1e-7$ for the CLIP encoder (keeping the normalization layers frozen)  and $1e-4$ for the shallow classifier. For ease of use, a classification loss (categorical cross-entropy) was used during this fine-tuning process. We fine-tuned the model for 30 epochs using the 2,920 classes not included in the test set.

\subsubsection{Experimental Results}
Before commenting on the results obtained we summarize the experimental setups:
\begin{itemize}
    \item \textbf{RN50 image features}: We compare the image features extracted with a ResNet-50 pretrained on ImageNet
    \item \textbf{CLIP image features}: We compare the image features extracted with the CLIP image encoder
    \item \textbf{CLIP class + text-to-image}: We perform a zero-shot classification of the query followed by a text-to-image retrieval using CLIP text and visual encoder
     \item \textbf{CLIP class + text-to-image + visual re-ranking}: We perform a visual re-ranking of the first 100 retrieved results after CLIP zero-shot classification and text-to-image retrieval
    \item \textbf{Oracle + CLIP text-to-image}: We perform only the text-to-image retrieval using the ground-truth class for the description
    \item \textbf{CLIP fine-tuned image features}: We compare the image features extracted with the CLIP image encoder after fine-tuning
\end{itemize}
\begin{table}[htb]
\centering
\begin{tabular}{c|| c}
\textbf{Experimental Setup} & \textbf{mAP} \\
\hline\hline
\textbf{RN50 image features} & 36.32 \\ \hline
\textbf{CLIP image features} &46.40 \\ \hline
\textbf{CLIP class + text-to-image} & 40.54 \\ \hline
\textbf{CLIP class + text-to-image + visual re-ranking} & 47.41 \\ \hline

\textbf{Oracle + CLIP text-to-image} & 54.21   \\ \hline
\textbf{CLIP fine-tuned image features} & \textbf{69.60}\\ 

\end{tabular}
\caption{Retrieval results on NoisyArt dataset using as queries the validation set and as index images the test set.}
\label{tab:noisyart-retrieval}
\end{table}

Table \ref{tab:noisyart-retrieval} summarizes the results of the experiments performed in the image retrieval setting previously described. It can be seen that CLIP visual features perform better than features extracted with a ResNet-50 pre-trained on ImageNet.
It is interesting to note that the re-ranking process makes the retrieval process performed by a zero-shot classification followed by a text-to-image retrieval operation more performing than the approach which uses only the visual features pre fine-tuning. This fact is obviously made possible by CLIP's good results in zero-shot classification illustrated in previous section.
It is also worth mentioning that using the ground-truth class and performing the text-to-image retrieval operation yields surprisingly good results: this confirms the goodness of CLIP in the text-to-image retrieval task. These results are even greater, by a significant margin, than those obtained using only visual features pre fine-tuning. This is probably due to the domain shift between validation and test set the visual features are more subject to.
Finally, we can see that CLIP fine-tuning was very successful, bringing a very significant performance boost and achieving better results than all other approaches.

\section{Conclusions}
In this paper we propose to use the zero-shot capabilities of CLIP in the artworks domain, showing how this approach can greatly improve over competing state-of-the-art approaches in the challenging NoisyArt dataset. Experiments show that in addition to zero-shot classification, the proposed approach can be used for content-based image retrieval, again outperforming by a large margin other competing approaches.
A benefit of using the proposed method is that it can be trained using very small datasets, thanks to the extensive pretraining of CLIP, and thus the method can be deployed also to be used on relatively small collections like those of small and medium-sized museums.

\paragraph{Acknowledgments}
This work was partially supported by the European Commission under European Horizon 2020 Programme, grant number 101004545 - ReInHerit.

\bibliographystyle{abbrvnat}
\bibliography{biblio}
%
%
%

\end{document}